# Clustering of Complex Networks and Community Detection Using Group Search Optimization

G. Kishore Kumar[†], Dr. V. K. Jayaraman

*Evolutionary Computing and Image Processing Group, Centre for Development of Advanced Computing(CDAC),
Pune University Campus, Ganeshkhind, Pune-411007, Maharashtra, India.
Email: kishorekumar2127@gmail.com, jayaramanv@cdac.in*

*Abstract*—Group Search Optimizer(GSO) is one of the best algorithms, and is very new in the field of Evolutionary Computing. It is very robust and efficient algorithm, which is inspired by animal searching behaviour. The paper describes an application of GSO to clustering of networks. We have tested GSO against five standard benchmark datasets and the algorithm is proven to be very competitive in terms of accuracy and convergence speed.

## I. Introduction

We are in the world of networks- our computers, mobiles and even we ourselves are part of complex social networks. The ability to measure the relationship between networks and different organizational attributes will enable us to create better working environments [1]. Once very least preferred research area like complex network analysis is now attracting much attention of mathematicians, computer scientists and even physicists [2]. Since it is at the amateur level, many aspects of the networks are not clearly understood, one needs to explore all the properties of the network to understand the dynamics of these networks.

By the very nature, these networks are complex to understand and formalise. Few empirical properties of networks are defined in the course of time, with the help of graph theoretical and statistical physics aspects like Degree distribution, Clustering coefficient, Community structure, Entropy etc [2]. Effort has to be placed in the direction of identification and integration of very crucial, sensitive and mutually consistent properties of the networks. One of those identifiedand well-studied properties is community structure. The structure detection is closely connected to graph partitioning in theory ofgraphs [19]. One can getbasic understanding of these aspects referring to Newman [2-4].

Newman and Girvan have proposed a measure of quality of a particular division of network, which is now famous Newman's modularity [2]. A particular network can be divided in to K communities, and this can be represented by a K×K symmetric matrix, we call it as modules matrix (e), $e_{ij}$ is the fraction of all network edges that link vertices in the group from i to j. Modularity measure is as follows:

$$Q = \sum(e_{ii} - a_i^2) = Tr(e) - \|e^2\| \qquad (1)$$

where $\|x\|$ is the sum of matrix x elements. Q is the best measure of community structure. Values of Q that are close to 1 represent better community structure. Since community structure identification is a graph partitioning problem, which is in turn an optimization problem, we need effective algorithms to tackle this problem. The potential candidates are evolutionary algorithms [5].

In recent years, various animal behaviour-inspired algorithms, which are known as swarm intelligence(SI) algorithms are coming up [7]. Most popular ones are Ant Colony Optimization [8] and Particle Swarm Optimizer [9]. One of the most recent entries in to SI algorithms family is GSO, proposed by He and Wu, based on animals foraging behaviour, which is primarily designed to handle continuous optimization problems [10]. Foraging tactics are very basic requirements, should be possessed by animals. Search for food is very common part of animal behaviour, which consumes most part of their life time. Efficacy of animal searching mechanism depends up on the group behaviour [10].

The GSO is a process of obtaining optimum solution in a search space, which is analogous to search for better clustering of a network to obtain best Newman's modularity. In the rest of the paper, we will first discuss the basics of GSO and later we get in to the details of implementation of the algorithm for network clustering. Even though many algorithms in literature addressed the clustering problem, still there is a need for efficient algorithms. Here in this paper we try toprove the superiority of the GSO over other available algorithms when applied to clustering problems.

## II. GROUP SEARCH OPTIMIZER

Group is the name given to population in GSO, and each individual is a member. Here we just want to give a few details of GSO, which are required to proceed with the analysis and application of algorithm. For more details on this algorithm, one needs to refer Hu [10], where it is very nicely and thoroughly explained. In GSO, the group consists of three types of members: producers, scroungers and rangers. Basically, there is only one producer at each searching bout and remaining members are scroungers and rangers (who perform random walk).

*a) Producers*

At each iteration, the candidate solution(Group member) conferring the best fitness value is chosen as producer. That member scans the search space for optimum position. Soon the producer will find a better position(Resources) with the best fitness value. If that position has a better resource than the current position, then producer will fly to this position or it will stay in current position and search for other optimal position. If the producer cannot find a better position after 'a'no. of iterations, it will retain back to its original position.

*b) Scroungers*

The Scroungers do take the best obtained by the producers, they always try to keep track of the best fitness values obtained by the producers. In case, the current producer can't find better fitness value, it will be replaced with one of those scroungers, which is next to producer, processing better fitness values. In case if a scrounger findsabetter optima in the course of time, in the next bout it will be made as a producer and all other members including the producer in the previous searching bout, will be kept to perform scrounging.

*c) Rangers*

The rest of the population comprises a group called rangers; they perform random walks in search of better resources. They always save the total population by not letting them get in to local optima. The potential of the algorithm lies in proper integration of different search strategies: systematic and random.

## III. IMPLEMENTATION DETAILS

Consider a population of 'm' members, whom we call as candidate solutions. Each candidate solution is an n-dimensional vector. Here 'n' represents number of nodes in the network. The $i^{th}$ element of the vector gives the details of cluster(module) number, where $i^{th}$ node is placed. TheFig.1 given below shows the details of the data structure used.

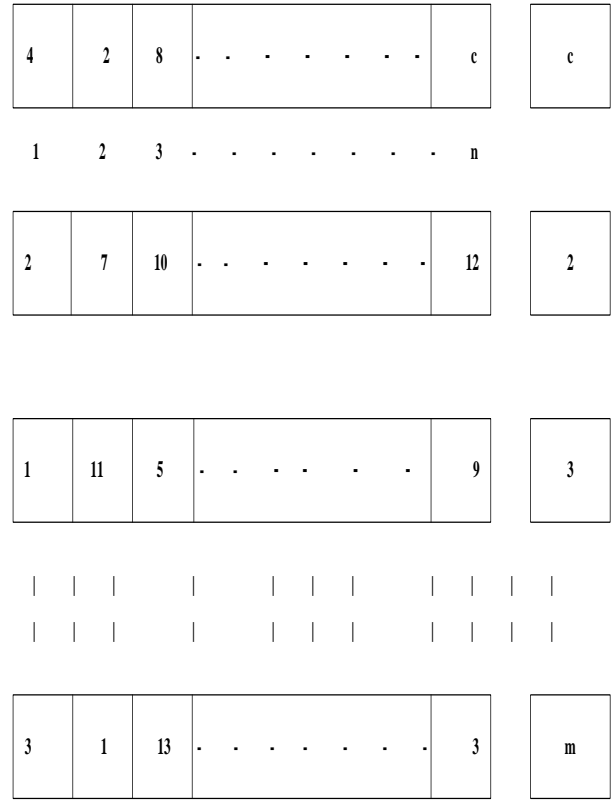

Fig. 1.Each member in a group is an n-dimensional vector, and each dimension possesses a cluster no. There are c no. of clusters in the complete network and m no of candidate solutions(members) in total population.

We randomly generate 'm' no of candidate solutions and each candidate solution in turn produces a Modules Matrix(MM), which gives the sum of all node connections(edges) between every two clusters. Using MM, Newman's modularity for each candidate solution is calculated. The candidate solution holding best fitness value(i.e., Newman's modularity value) is chosen as producer and remaining all candidate solutions are selected as scroungers and rangers.

Perform GSO for 'j' no of iterations. Basically, each candidate solution(member) corresponds to a different way of clustering the nodes.During each iteration, the members of the group try to attain new positions in accordance with GSO. After each iteration, fitness value of each candidate solution on its new positionis calculated. The current producer is replaced with the candidate solution which attains best fitness value. The best fitness value and corresponding candidate solution is recorded for further use. The Fig. 2 given below shows the flow chart for implementation of GSO.

## IV. EVALUATION

The code for the algorithm has been written in GNU Octave and tested on five social network datasets: Zachary karate club, Dolphins network, Jazz Musicians network, American football network and Les Miserables novel characters network.

| Algorithms | Modularity |
|---|---|
| Newman | 0.381 |
| DA | 0.419 |
| GN | 0.401 |
| EO | 0.424 |
| GSO | 0.613 |

Table 1: Results of GSO on karate club data

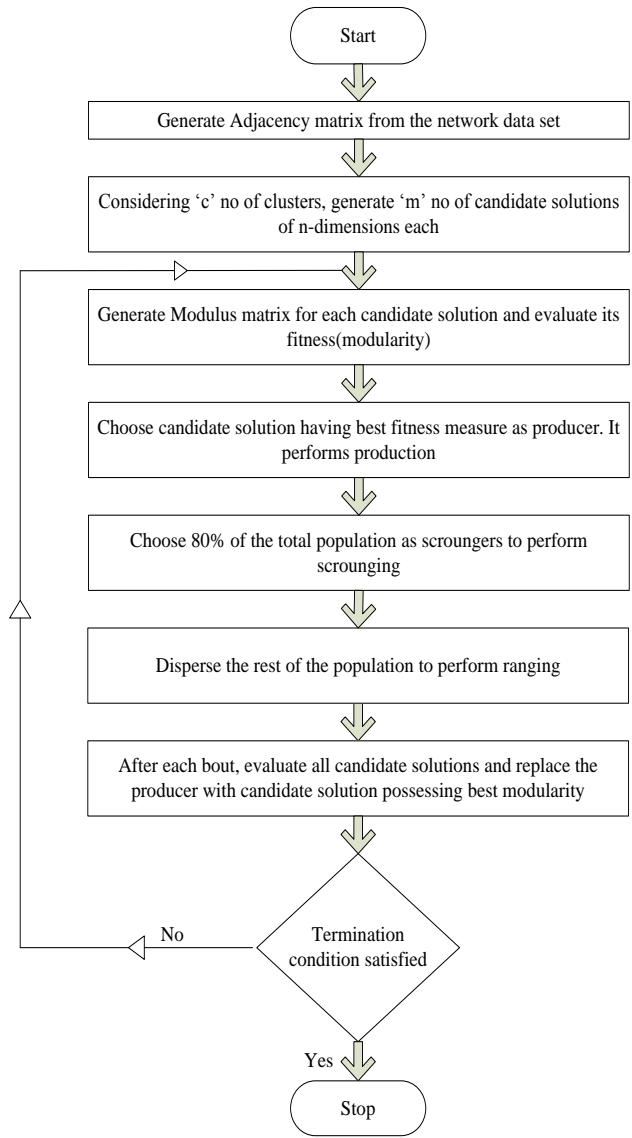

Fig. 2. Flow chart for the implementation of GSO.

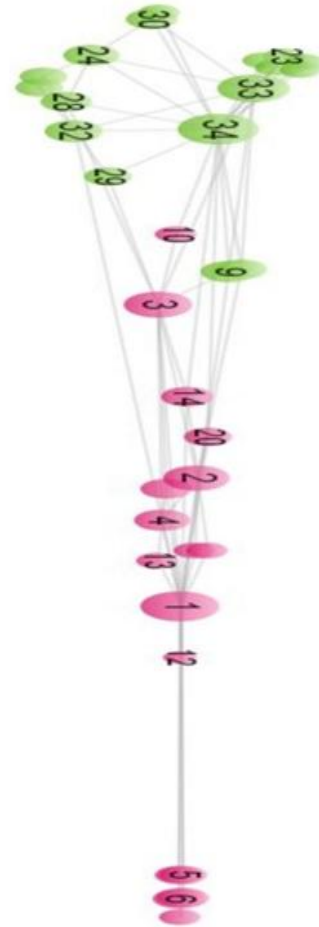

Fig.3. The friendships among individuals in the karate club network are represented. Nodes 1 and 33 representthe administrator and instructor, respectively. Green colour circles represent individuals who are joined with the club's administrator, and pink circles are individuals who joined hands with the instructor after fission.

*a) Zachary karate club network*

This is one of the well-studied network datasets, in which, relations between the members of karate clubs are described [11]. Prior to internal conflict, the club had a friends group of 34 people. Network has 34 nodes and 78 edges [21]. Table 1 shows the results of modularity obtained by GN [12], Newman [13], DA [14], EO [15] and GSO. One can easily notice how GSO is much superior to other methods.

*b) Jazz musicians' network*

This is collected from the Red Hat Jazz archive [18]. The network gives the details of collaborations between early Jazz musicians of Gleiser and Danon. It has 196 nodes and 2742 edges [21]. Table 3 compares the modularity obtained by GN [12], Newman [13], DA [14] EO [15] and GSO.

| Algorithms | Modularity |
|---|---|
| Newman | 0.438 |
| DA | 0.445 |
| GN | 0.405 |
| EO | 0.468 |
| GSO | 0.520 |

Table 2: Results of GSO on Jazz musicians' network

*c) American football network*

This network was drawn from the schedule of games played between 115 NCAA divisions. I-A American college football teams in the year 2000 [12]. It has 115 nodes and 615 edges [21]. Table 4 shows the results of modularity obtained by Newman [20], spectral [20] and GSO.

| Algorithms | Modularity |
|---|---|
| Newman | 0.556 |
| Spectral-2 | 0.553 |
| GSO | 0.604 |

Table 3: Results of GSO on American football network

*d) Les Miserables*

This network gives the details of interactions between major characters in Victor Hugo's novel of crime and redemption in post-restoration France [22]. The network has 77 nodes and 254 edges [21]. Table 5 compares the modularity values obtained by Newman [20], CNM heuristic [23] and GSO.

| Algorithms | Modularity |
|---|---|
| Newman | 0.540 |
| CNM heuristic | 0.500 |
| GSO | 0.630 |

Table 4: Results of GSO on Les Miserables characters data

*e) Dolphins network*

An undirected social network of frequent associations between 62 dolphins in a community listing of doubtful sound, New Zealand [16,17]. Network has 62 nodes and 159 edges [21]. Table 2 show the results of GSO compared with modularity obtained by Newman [13], GN [12], EO [15] and GSO.

| Algorithms | Modularity |
|---|---|
| Newman | 0.52 |
| GN | 0.52 |
| EO | 0.53 |
| GSO | 0.623 |

Table 5: Results of GSO on Dolphins network data

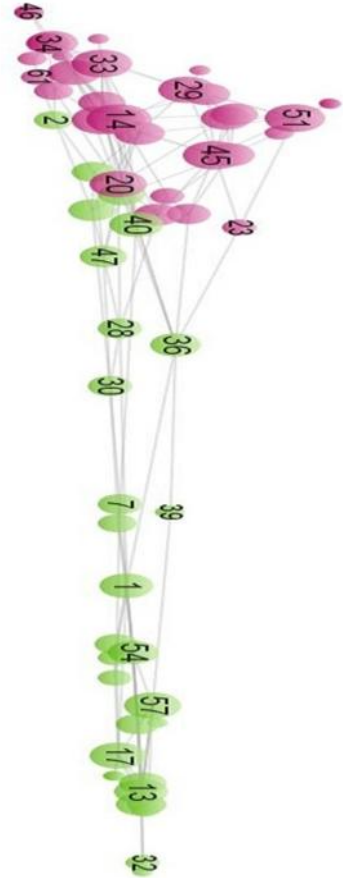

Fig.4. The social structure of the bottlenose dolphin community. The network is divided in to two groups as shown above. Pink circles represent one group and green circles represent the other.

V. ACKNOWLEDGMENT

The authors thank the Centre for Development of Advanced Computing(CDAC), Pune, India, for providing computational facilities, software and great working environment. Thanks also to M.E.J. Newman for keeping datasets available to public for free of cost, and to Dr.Sanjay Kadam, Shameek Ghosh, Kalpesh, B.Rajasekhar, Vinay Nair and Utkarsh Gupta for helpful discussions.